\documentclass[aip, rsi,amsmath,amssymb,reprint]{revtex4-2}
\usepackage{xcolor}

\usepackage{graphicx}% Include figure files
\usepackage{dcolumn}% Align table columns on decimal position
\usepackage{amsmath}
\usepackage{subfig}
\usepackage{bbm}
\usepackage{bm}% bold math
\usepackage{float}
\usepackage{placeins}

\usepackage{amsmath}

\begin{document}

\title[]{Machine Learning for Detection of 3D Features using sparse X-ray data}

%\color{lightgray}
\author{Bradley T. Wolfe} 
\affiliation{Los Alamos National Laboratory, Los Alamos, New Mexico 87545, US}
\author{Michael J. Falato}  
\affiliation{Los Alamos National Laboratory, Los Alamos, New Mexico 87545, US}
\author{Xinhua Zhang}
\affiliation{Los Alamos National Laboratory, Los Alamos, New Mexico 87545, US}
\author{Nga T. T. Nguyen-Fotiadis} 
\affiliation{Los Alamos National Laboratory, Los Alamos, New Mexico 87545, US}
\author{J.P. Sauppe} 
\affiliation{Los Alamos National Laboratory, Los Alamos, New Mexico 87545, US}
\author{P. M. Kozlowski}
\affiliation{Los Alamos National Laboratory, Los Alamos, New Mexico 87545, US}
\author{P. A. Keiter} 
\affiliation{Los Alamos National Laboratory, Los Alamos, New Mexico 87545, US}
\author{R. E. Reinovsky} 
\affiliation{Los Alamos National Laboratory, Los Alamos, New Mexico 87545, US}
\author{S. A. Batha} 
\affiliation{Los Alamos National Laboratory, Los Alamos, New Mexico 87545, US}
\author{Zhehui Wang} 
\affiliation{Los Alamos National Laboratory, Los Alamos, New Mexico 87545, US}
\date{\today}

\begin{abstract}
    %\color{lightgray}
    In many inertial confinement fusion experiments, the neutron yield and other parameters cannot be completely accounted for with one and two dimensional models.
    This discrepancy suggests that there are three dimensional effects which may be significant.
    Sources of these effects include defects in the shells and shell interfaces, the fill tube of the capsule, and the joint feature in double shell targets.
    Due to their ability to penetrate materials, X-rays are used to capture the internal structure of objects.
    Methods such as Computational Tomography use X-ray radiographs from hundreds of projections in order to reconstruct a three dimensional model of the object.
    In experimental environments, such as the National Ignition Facility and Omega-60, the availability of these views is scarce and in many cases only consist of a single line of sight.
    Mathematical reconstruction of a 3D object from sparse views is an ill-posed inverse problem. 
    These types of problems are typically solved by utilizing prior information.
    Neural networks have been used for the task of 3D reconstruction as they are capable of encoding and leveraging this prior information. 
    We utilize half a dozen different convolutional neural networks to produce different 3D representations of ICF implosions from the experimental data.
    We utilize deep supervision to train a neural network to produce high resolution reconstructions.
    We use these representations to track 3D features of the capsules such as the ablator, inner shell, and the joint between shell hemispheres. 
    Machine learning, supplemented by different priors, is a promising method for 3D reconstructions in ICF and X-ray radiography in general.
\end{abstract}

\keywords{Machine Learning, Neural Networks, 3D Reconstruction, X-Ray Imaging, radiography,  x-ray imaging}
\maketitle
\section{Introduction}
Inertial Confinement Fusion (ICF) experiments contain three dimensional effects that produce results which deviate from low dimensional simulation. \
Double shell ICF targets\cite{doi:10.1063/1.5042478} consist of an ablator (outer shell), a pusher (inner shell), a mid-Z tamper (outer layer of the pusher), and a foam cushion between the shells.
The inner shell is then filled with $D_2$ or a D-T mix.
During implosions the ablator is driven towards the pusher, which is used to compress the fuel to fusible conditions.
Understanding the effects of asymetries from the ablator on the pusher is important, as these asymetries affect the compression of the fuel.
The asymmetry in the fuel compression can cause loss in neutron yield from fusion processes.
Ablator asymetries can stem from asymmetric drive, presence of joint features, and presence of fill tubes \cite{Kline_2019,Springer_2018,doi:10.1063/1.4959117}.
While X-ray radiography\cite{10.1117/12.513761} provides a platform for resolving these features, multiple views are typically not available and thus this method provides a two dimensional view of the object.
Neural networks have been shown to be able to construct 3D models from images at a single view.\cite{wolfe2021}  
These models however, are difficult to interpret and act as a black box.
In this paper, we compare the reconstructions from multiple neural network architectures, improve upon these results, and introduce high resolution reconstructions.

We use the following process to reconstruct double shells from a single view.
We create a model of a series of shells and ray trace the projection of the object.
We utilize generative adversarial networks to transfer noise characteristics from experimental data to the priojection.
We train neural networks to map an image to the density map of the object.
Given the predicted density map, we extract a mesh and fit a spherical harmonic representation.
\section{Synthetic Data Generation}
\label{Synthetic Data Generation}
\begin{figure}
  \includegraphics[width=.48\textwidth]{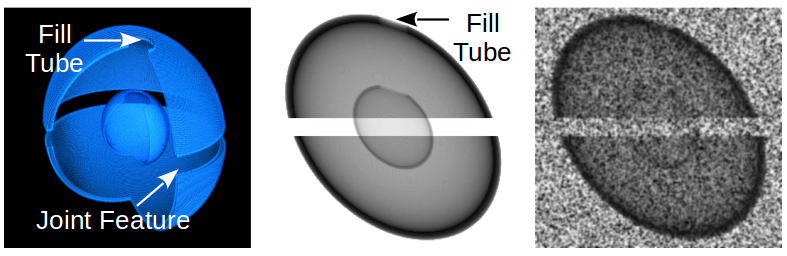}
  \caption{An example of synthetic data. 
           (Left) An object generated at a $256^3$ resolution. 
           (Middle) The object projected using the beer-lambert inversion.
           (Right) Noise and blur added to the object.}
  \label{syn}
\end{figure}

In order to train neural networks images paired with groud truth (or known) reconstructions are required.
Due to the scarcity of ICF data and the unavailability of a ground truth object, a mathematical model is used to produce synthetic data to assist with training the neural network. 
For a monochromatic X-ray source, the transmission of X-rays through a material is given by the Beer-Lambert Law
\begin{equation}
T := e^{-\int_L\mu(x,y,z)dl}
\end{equation}
where $\mu$ is the linear attenuation coefficient\cite{osti_6016002} and $L$ is a line between the X-Ray source and the object.
These integrals are calculated using a ray tracing algorithm provied by the python package TIGRE\cite{Biguri_2016}.
The linear attenuation of object is desribed by the mass attenuation which is a property of the material and the density of the the object.
The object is described with parameters $C$ which define a function $\rho(x,y,z;\mathbf{C})$, where $\rho(x, y, z)$ is the density of the object at position $(x, y, z)$. 
For an estimate, objects are treated as shells with constant densities. 
This allows for the model to be broken down into models that describe each shell:
\begin{equation}
\rho(x,y,z;\mathbf{C}) = \sum_i \rho_{i}\mathbbm{1}_{i}(x, y, z; \mathbf{C}_i)
\end{equation}
Where each $\mathbbm{1}_{i}$ is an indicator function for the $i^{th}$ shell that is 1 at locations where the shell is located and 0 otherwise. 
The $\mathbf{C}_i$ are the parameters that describe the $i-th$ shell. 
This model can be extended to any number of shells and allows for a variety of object geometries.
To produce objects without spherical symmetry, ellipsoidal data is generated using a spherically symmetric shell model and applying a shear transformation.
The general shear transformation is given by
\begin{equation}
S = 
\begin{bmatrix}
1       & s_{xy} & s_{xz} & 0\\
s_{yx}  & 1      & s_{yz} & 0\\
s_{zx}  & s_{zy} & 1      & 0\\
0       & 0      & 0      & 1\\
\end{bmatrix}
\end{equation}
where the matrix acts on a vector $[x, y, z, 1]$.
Figure \ref{syn} shows the data produced using TIGRE.

Another method for generating asymmetry is Legendre polynomials.
When utilizing Legendre polynomials the shell interface is decribed by
\begin{equation}
  R(\theta, \phi) = \sum_{n=0}^N a_nP_n(\cos (\theta))
\end{equation}
where the interface has azimuthal symmetry.
We generate shells with low mode asymetries $(N=4)$.

\subsection{Joint Feature and Fill Tube}
Modelling the object as shells with constant densities ignores some common features found in double shell images, such as the joint feature and fill tube.
The joint feature is modelled by introducing a gap into the model which splits the outer shell into two hemispheres.
This adds a parameter to the model $s$ which is the size of the joint feature.
Given that the outer shell is defined by $\mathbbm{1}_{shell}(x, y, z; \mathbf{C})$, where the outer surface is $r_{out} = r_\mathbf{C}(\theta, \phi) + \frac{\tau}{2}$ and and the inner surface is $r_{in} = r_\mathbf{C}(\theta, \phi) - \frac{\tau}{2}$, the joint can be described by $s,\ r_\mathbf{C},\ r_{in},\ and\ r_{out}$. 
The joint is given as 
\begin{equation}	
\mathbbm{1}_{joint}(x, y, z;\mathbf{C}) = \begin{cases}
1 & r_{in} \leq r_\mathbf{C} \leq r_{out} \\
  &	and -\frac{s}{2}  \leq z \leq \frac{s}{2} \\
0 & otherwise
\end{cases}
\end{equation} 
The new shell model is then described as $\mathbbm{1}'_{shell} = \mathbbm{1}_{shell} - \mathbbm{1}_{joint}$.

The fill tube is modeled by introducing a cylinder into the model.
This adds a parameter to the model $r$ which is the radius of the fill tube.
The fill tube is given as
\begin{equation}
\mathbbm{1}_{fill tube}(x,y,z;\mathbf{C}) = \begin{cases}
1 & z > 0\ and\ \sqrt{x^2 + y^2} < r \\
0 & otherwise
\end{cases}
\end{equation}

\section{Texture Synthesis using Generative Adversarial Networks}
\begin{figure}  
  \includegraphics[clip,width=.48\textwidth]{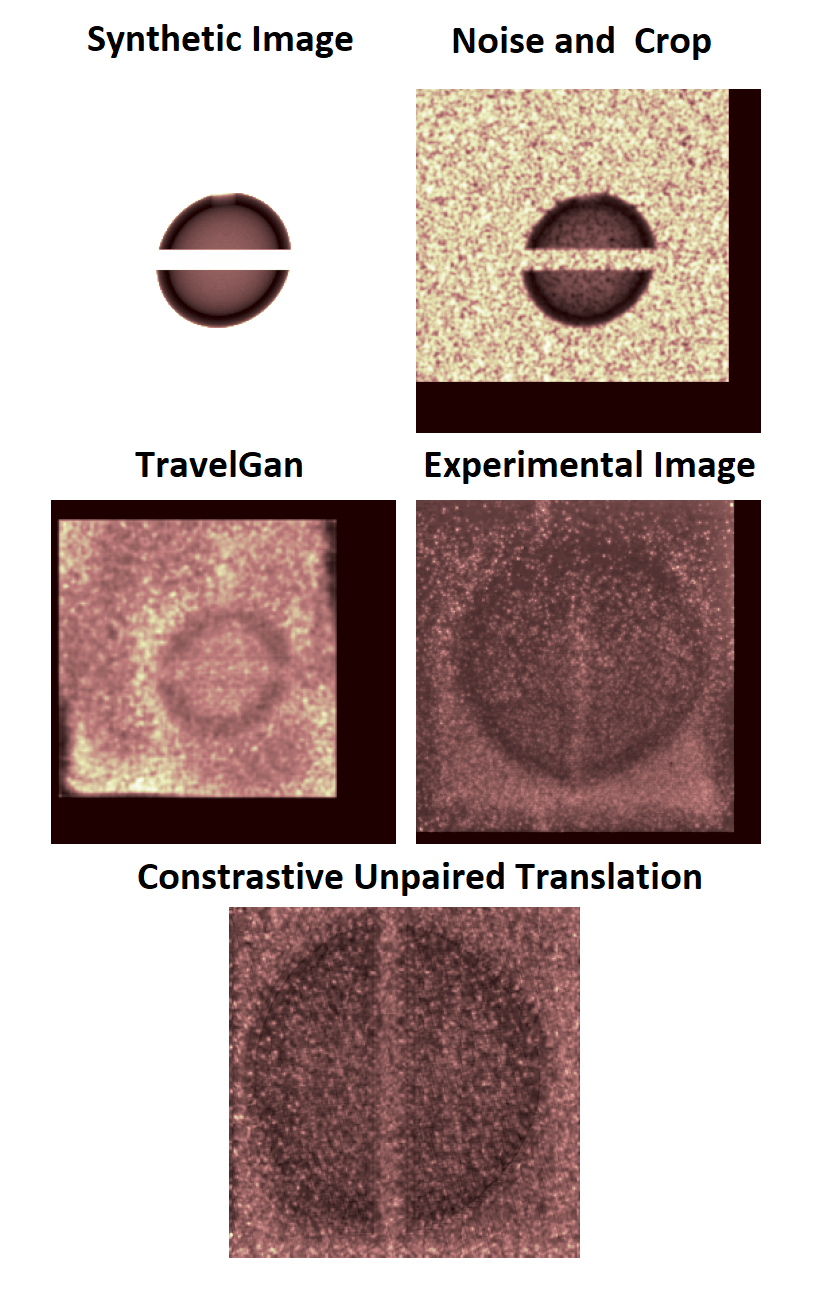}
  \caption{Application of noise to synthetic radiographs. 
  Since TraVeLGAN views the entire image it is capable of applying transformations such as cropping. 
  CUT produces noise dependent on the region of the image.}
  \label{GAN}
\end{figure}

In order for mappings from synthetic data to 3D volumes to apply to experimental data, the synthetic data must have similar noise characteristics to the experimental data.
In ICF environments there are multiple sources of noise, such as, defects in filters an blast shields, ccd noises including dark current and quantization noise, and the stochastic nature of gain of the microchannel plate.
Overall this noise is neither additive or follows typical noise distributions, so instead of designing a forward model, we utilize  a generative model to produce realistic noise for our synthetic radiographs.
We use two generative adversarial networks, Transformation Vector Learning GAN(TraVeLGAN)\cite{TraVeLGAN} and Contrastive Unpaired Translation(CUT)\cite{park2020contrastive}, to fit a model for generating the experimental-like radiographs.
In both cases the networks consist of a generator network and a discriminator network.\cite{https://doi.org/10.48550/arxiv.1406.2661} 
The generator network takes in a synthetic image as an input and outputs the image with the noise model applied.
The discriminator network determines whether a provided image was produced by the generator or is an actual experimental image.
The generator is optimized so that the probability that the discriminator correctly identifies the image type is minimized.
The discriminator is optimized so the probability that it correctly identifies the type of image is maximized.
Both CUT and TraVeLGAN also have a network that measures similarity of the synthetic image and the generated image.
This network is used to inform the generator so that generated images possess common features to the synthetic images.
TraVeLGAN uses a Siamese neural network\cite{Koch2015SiameseNN,10.5555/2987189.2987282} which compares entire images, whereas CUT utilizes the encoder portion of the generator to compare the encodings of patches of images. 
This is used to preserve features such as shell boundaries, fill tubes, and joint features.
Figure \ref{GAN} shows the noise models produced from CUT and TraVeLGAN when applied to TIGRE data.

For training our models, we produce a dataset of 10,000 pairs for TraVeLGAN data and 2000 pairs for CUT data.
For validation during training we use a set of 40 pairs.

\section{3D Reconstruction Models}
\label{methods}
\subsection{Encoder-Decoder Model} 
\begin{figure}
  \includegraphics[clip,width=.48\textwidth]{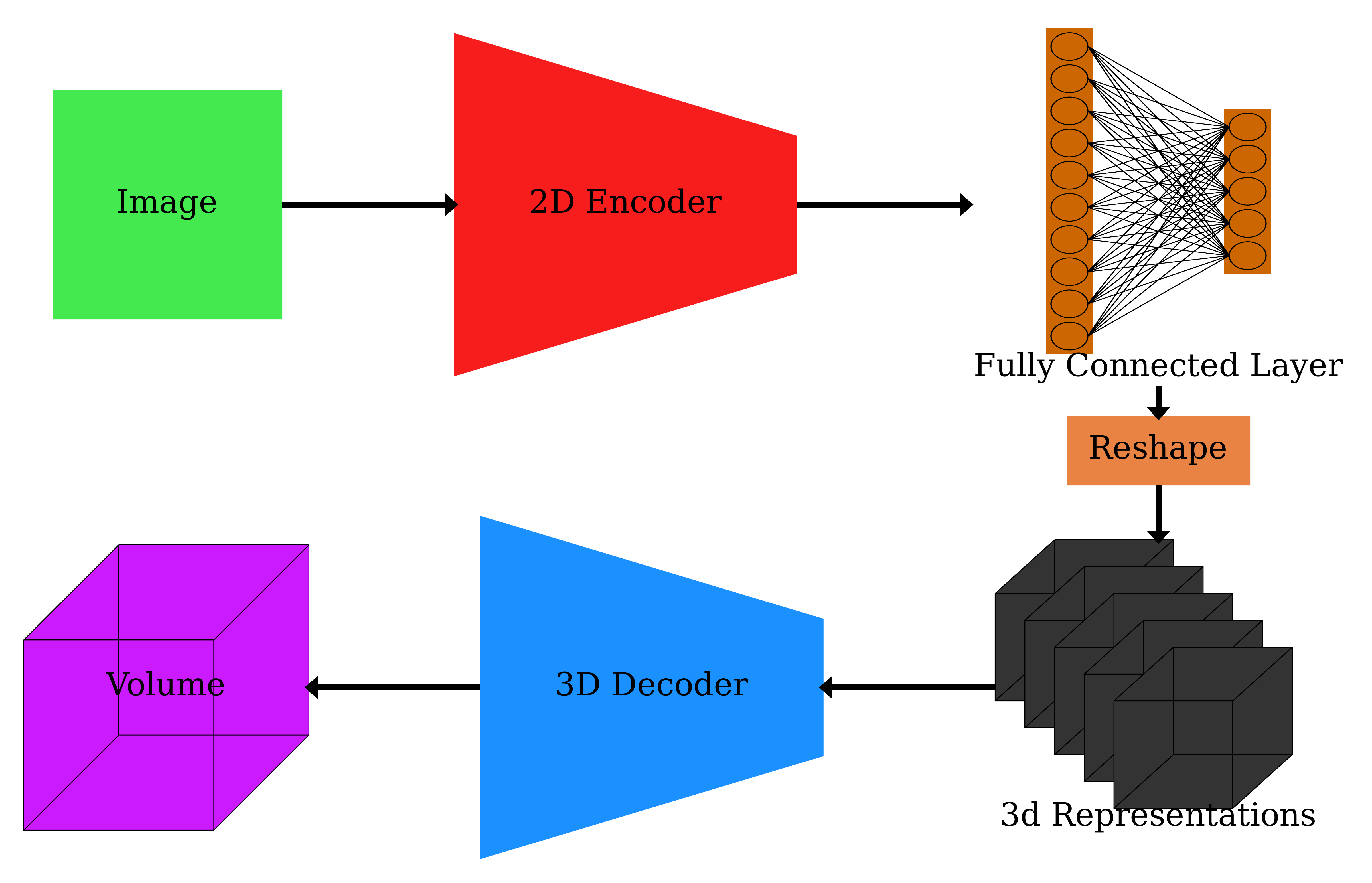}
  \caption{The architecture of a 3D volumetric reconstruction neural network. 
  The output of the encoder is flattened and passed through a fully connected layer.
  The output of this layer is reshaped to form the 3D representations. 
  In some networks the 2D representations from the encoder are directly reshaped into 3D representations without using a fully connected layer.}
  \label{e2dArch}
\end{figure}

In 3D reconstruction, neural network models generally consist of an encoder and a decoder (shown in Figure \ref{e2dArch}). 
The encoder utilizes 2D strided convolutional filters to generate an internal representation of the image.
The decoder uses 3D transposed convolutions or upsampling with 3D convolutions to reproduce the 3D model from the internal representation.
These models can be either fully convolutional\cite{https://doi.org/10.48550/arxiv.1411.4038} by reshaping the output of the encoder or include fully connected layers before reshaping when passing the output to the decoder.

The model can be trained by minimizing the $L_2$ loss,
\begin{equation}
\mathcal{L} = \sum_{ijk} (Y_{ijk} - f(X)_{ijk})^2
\end{equation}
where Y is the ground truth volume, X is the input projection, f is the function to be optimized, and $ijk$ is a voxel coordinate. 
If the outputs are scaled to [0, 1] the binary cross entropy can be used.
\begin{equation}
\mathcal{L} = \sum_{ijk} [-Y_{ijk}\log(f(X)_{ijk}) + (1-Y_{ijk})\log(Y_{ijk}-f(X)_{ijk})]
\end{equation}

\subsection{Deep Supervision}
\begin{figure}
  
  \includegraphics[clip,width=.48\textwidth]{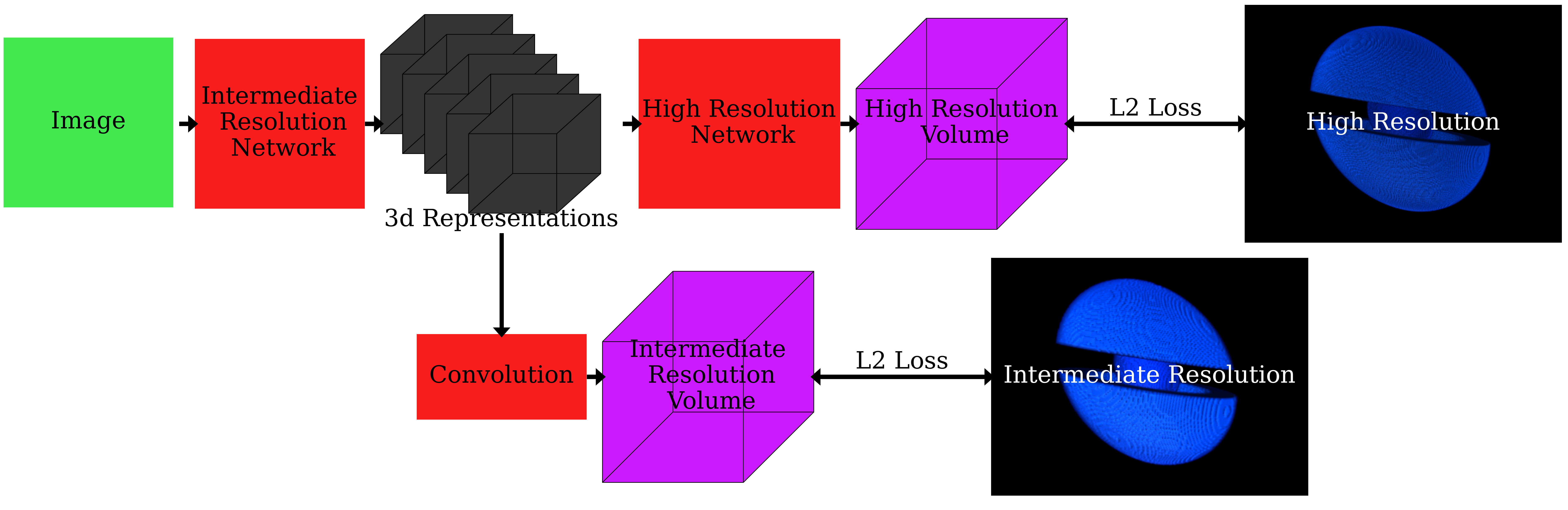}%.96
  \caption{An example 3D reconstruction architecture that is trained using deep supervision with one intermediate representation. 
  Note that the model is simultaneously fit to the reconstruction at multiple resolutions.}
  \label{deepsupervision}
\end{figure}

Low resolution reconstruction models capture large scale aspects of the 3D object.
In order to generate high resolution reconstructions that capture small scale features, we add 3D transposed convolutional layers to upscale the representations.
Adding more layers to the network and Increasing the output resolution makes training more difficult.
Since low resolution models are easier to train and both low and high resolution models share similar large scale features, we leverage a low resolution model by utilizing utilize deep supervision\cite{lee2015deeply}.
Deep supervision is the process of fitting intermediate outputs of the network to the ground truth.
This mitigates vanishing gradients which are produced by saturation of activation functions and can speed training and improve performance of the network.
We use the MVD base model and supervise at multiple resolutions of $64^3$, $128^3$, and $256^3$. 
In order to train this network, we downsample the ground truth data and take a weighted average of the loss between the model and the downsampled ground truth:
\begin{equation}
\mathcal{L} = \sum_i \lambda_i \mathcal{L}(f(X; R_i), D_i(Y))
\end{equation}
where $D_i$ is a function that downsamples the data and $f(X; R_i)$ is the output of the network at a particular resolution $R_i$.
Figure \ref{deepsupervision} shows a diagram of the the deep supervision process.

\section{3D Reconstruction}
\label{Reconstruction}
\begin{figure}
  \centering
  \includegraphics[width=.48\textwidth]{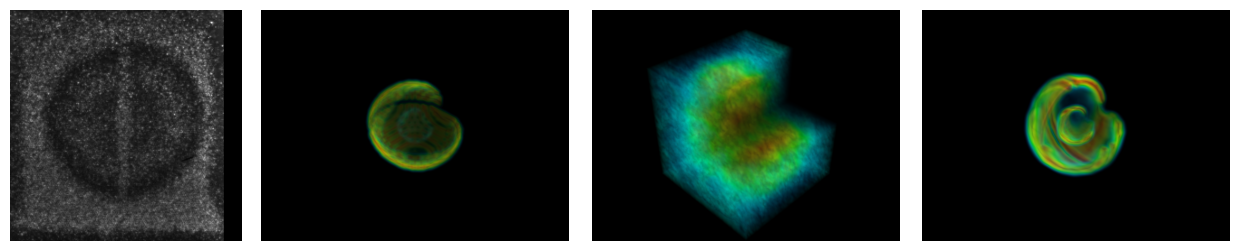}
  \includegraphics[width=.48\textwidth]{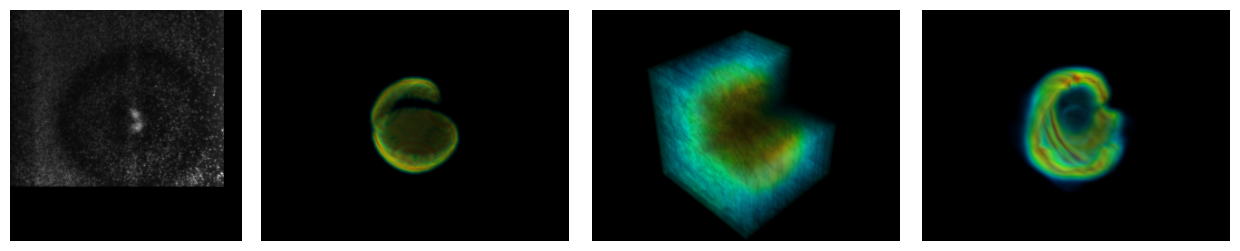}
  \caption{(Left to Right) Input experimental image, AttSets reconstruction, TBN reconstruction, MVD reconstruction.
  The MVD and AttSets methods generate clearer reconstructions than the TBN reconstructions.}
  \label{3netexp}
\end{figure}

\begin{figure}
  \centering
  \includegraphics[width=.48\textwidth]{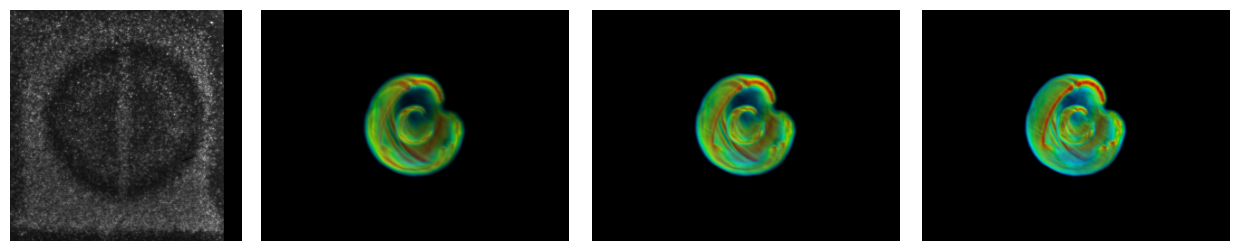}
  \includegraphics[width=.48\textwidth]{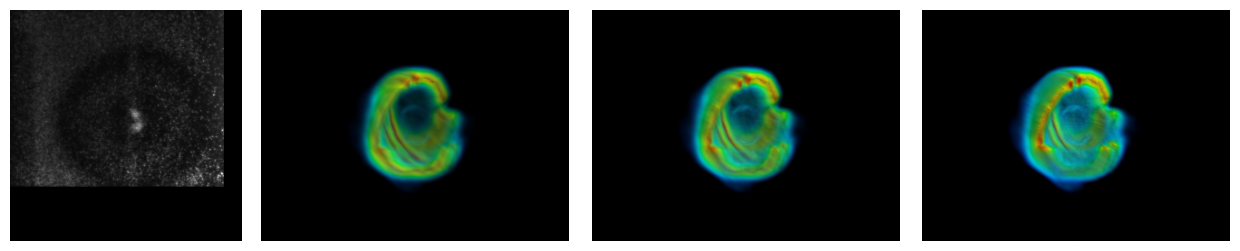}
  \caption{(Left to Right) Input experimental image, $64^3$ resolution, $128^3$ resolution, $256^3$ resolution.
  Though the reconstructions are mostly conistent, changing the generated resolution changes details in the reconstructions.}
  \label{3resexp}
\end{figure}

\begin{figure}
  \centering
  \includegraphics[width=.48\textwidth]{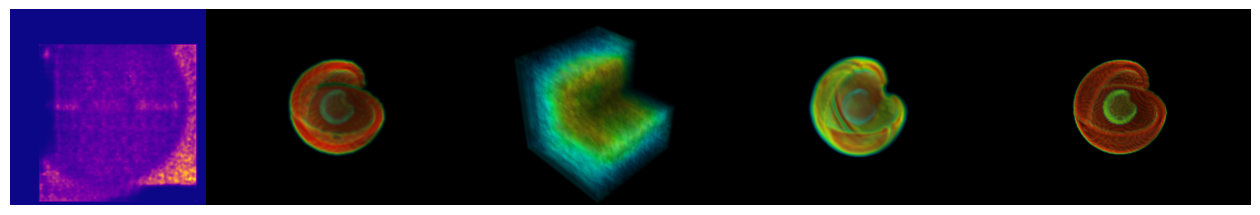}
  \caption{(Left to Right) Input TraVeLGAN image. AttSets Reconsturction. TBN reconstruction. MVD reconstruction. Ground Truth 3D Model.
            Even on synthetic data the TBN based model is less clear than the other models.}
  \label{3netsyn}
\end{figure}

\begin{figure}
  \centering
  \includegraphics[width=.48\textwidth]{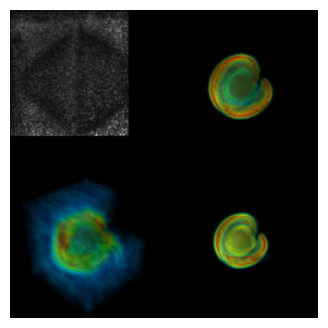}
  \caption{(Top Left) Experimental image used in reconstruction (Top Right) Reconstruction from AttSets (Bottom Left) Reconstruction from TBN (Bottom Right) Reconstruction from MVD.
            While the other architectures already produce clearer reconstructions than TBN, TBN trained with data produced using CUT produces clearer reconstructions than  when trained using data produced using TraVeLGAN.}
  \label{cutexp}
\end{figure}

We utilize three neural networks for 3D reconstruction.
These include the modified encoder of Transformable Bottelneck Networks\cite{olszewski2019transformable}, the R2N2\cite{choy20163d} encoder from AttSets\cite{Yang2020}, and the low resolution reconstruction network from MVD\cite{smith2018multi} modified to use deep supervision.
AttSets and MVD are networks commonly used in the task of 3D occupancy reconstruction.
TBN is used to generate a new view from a set of input views in a task called novel view synthesis by building a 3D occupancy representation.
MVD and AttSets were both trained using a cross entropy loss and TBN was trained using an $L_2$ loss.
Prior to passing the experimental data to the networks, we apply a pseudo-flat field correction\cite{XRIPL} to remove out large intensity gradients of the image.
These networks were trained on two synthetic data models, ellipsoidal objects with noise models produced from Legendre polynomial based objects with noise models produced from CUT.
Figures \ref{3netexp} and \ref{cutexp} show reconstructions of experimental images using different networks trained on TraVeLGAN and CUT repectively.
Figure \ref{3netsyn} shows reconstructions of a synthetic TraVeLGAN image. 
We find that the MVD and AttSets networks can fit the data while the TBN network produces a reconstruction with significant background.
We also find the the the reconstructions produced using MVD trained with the CUT model produces the highest quatlity reconstructions.
Figure \ref{3resexp} shows reconstructions generated by MVD when utilizing deep supervision. %and \ref{3ressyn}
This shows the capabiblity of producing high resolution 3D reconstructions with low background.
Table \ref{table:Error} shows the mean squared error for reconstructions of synthetic models using CUT based data using 2000 test images.
MVD and AttSets outperform TBN, which corresponds with the cloudy vizualiztions. 
Increasing the reconstruction resolution does not have a negative impact on the error, which is shown by the MVD errors.
Although AttSets produces a smaller reconsruction error than MVD it is limited to a much lower resolution.

\begin{table}[]
  \begin{tabular}{|l|l|l|l|l|l|}
  \hline
  Network&AttSets(32)&TBN(64)&MVD(64)&MVD(128)&MVD(256)\\ \hline
  MSE&0.01289&0.05148&0.03462&0.03462&0.03463 \\ \hline
  \end{tabular}
  \caption{Mean squared error of the networks on CUT based synthetic data with 2000 images. 
  The number in parenthesis is the resolution of the reconstruction.}
  \label{table:Error}
\end{table}

%\begin{figure} 
%  \centering
%  \includegraphics[width=.48\textwidth]{figures/3res/17.png}
%  \caption{(Left to Right) Input experimental image, $64^3$ resolution, $128^3$ resolution, $256^3$ resolution.
%            The top row is are the predicted models and the bottom row are the ground truth models.}
%  \label{3ressyn}
%\end{figure}

\section{3D Feature Extraction}
Since MVD with deep supervision can produce high resolution reconstructions with low background, we are able to extract features of the 3D object.
In order to extract features from the image such as the shell boundaries, we utilize marching cubes\cite{lorensen1987marching} to produce a mesh from the object. 
The center of the object is then calculated using:
\begin{equation}
\vec{r}_{center} = \frac{\int_\Omega \vec{r} dS}{\int_\Omega dS}
\end{equation}
The surface elements in this case are the triangular faces of the mesh. 
The $\vec{r}$ is given by the center of the triangular face.

We apply a clustering algorithm, DBSCAN\cite{Ester1996ADA}, on the vertices of the mesh to separate both shells and interior and exterior surfaces.
Then for a surface, we find the radial distance $(r)$, polar angle $(\theta)$, and azimuthal angle $(\phi)$.
In order to evaluate low-mode 3D asymetries of the object, we fit the $(r,\theta,\phi)$ point cloud with real-valued spherical harmonics:

\begin{equation}
  r(\theta, \phi) = \sum_{n=0}^N\sum_{m=-n}^n c_{nm}Y_{nm}(\theta, \phi)
\end{equation}

\begin{figure}
  \centering
  \includegraphics[width=.48\textwidth]{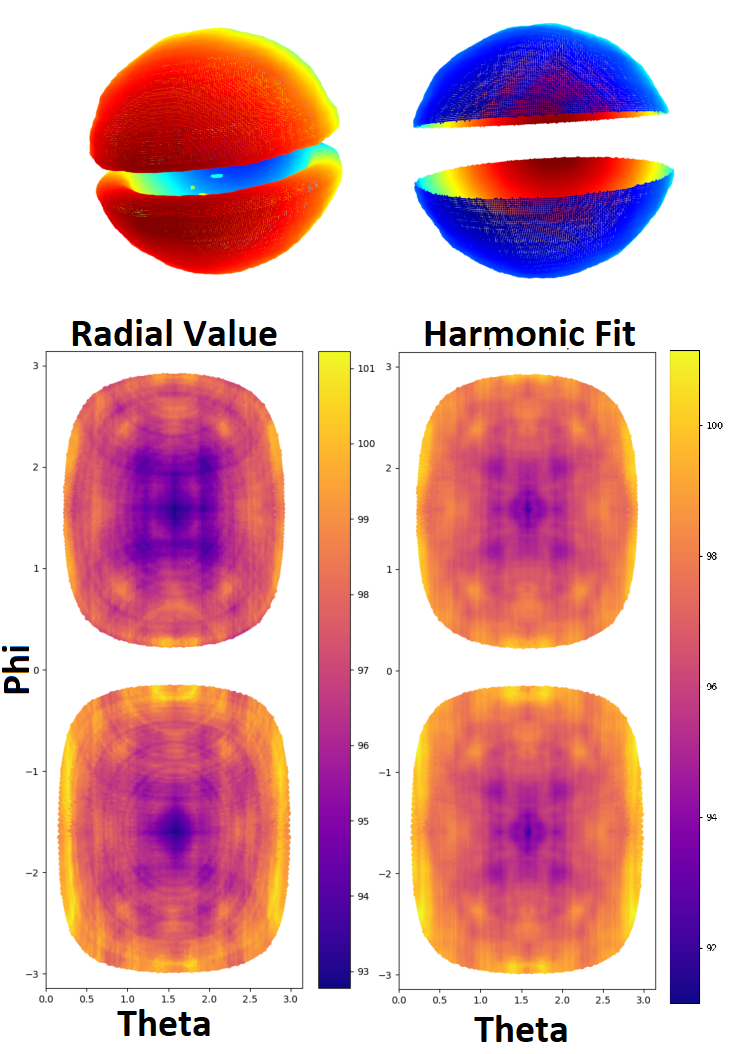}
  \caption{Using marching cubes we extract a mesh from the experimental data (Top Left).
           We then select part of the shell surface so that a fit can be applied (Bottom Left).
           We are then able to fit a spherical harmonic series to the data $(N=10)$ (Bottom Right).}
  \label{fitting}
\end{figure}

Figure \ref{fitting} shows the process of extracting a shell surface and fitting the radial data.
We find that this method works well for clear reconstructions such as MVD, but for blurry reconstructions produced by TBN, the mesh is unable to be extracted.
This method also works better for the higher resolution objects since DBSCAN is density based and the results are less sensitive to parameter selection. 

\section{Conclusion}
We compared multiple neural networks for 3D reconstruction of ICF radiographs. 
We show that generative models can be used for improving synthetic radiographs and producing images that are similar to experiment.
We show that by utilizing deep supervision we can achieve higher resolutions of reconstructions.
By combining these techniques we can achieve clear high resolution 3D reconstructions where a mesh of surfaces can be extracted.
These meshes can be fit using representations such as spherical harmonics which allow for characterization of 3D asymmetry of the object.
The joint features from images are present in the reconstructions which enable 3D analysis of the joint feature. 
Futher work on experimental validation of the generative and reconstruction methods will be required.
\section{Acknowledgment}
This work was supported by the U.S. Department of Energy through the Los Alamos National Laboratory. Los Alamos National Laboratory is operated by Triad National Security, LLC, for the National Nuclear Security Administration of U.S. Department of Energy (Contract No. 89233218CNA000001).

\bibliography{main}
\end{document}